%% file: cvpr-2026/main.tex
\documentclass[10pt,twocolumn,letterpaper]{article}

\usepackage[pagenumbers]{cvpr}              % To produce the CAMERA-READY version

\usepackage{amsmath}
\usepackage{amssymb}
\usepackage{mathtools}
\usepackage{amsthm}
\usepackage[utf8]{inputenc} % allow utf-8 input
\usepackage[T1]{fontenc}    % use 8-bit T1 fonts
\usepackage{url}            % simple URL typesetting
\usepackage{graphicx}
\usepackage{booktabs}       % professional-quality tables
\usepackage{multicol}
\usepackage{multirow}
\usepackage{amsfonts}       % blackboard math symbols
\usepackage{nicefrac}       % compact symbols for 1/2, etc.
\usepackage{microtype}      % microtypography
\usepackage{xcolor}         % colors
\usepackage{amsmath}
\usepackage[font=footnotesize]{caption}
\usepackage{subcaption}
\usepackage{scalerel}
\usepackage{svg}
\usepackage{placeins}
\input{cvpr-2026/preamble}
\definecolor{cvprblue}{rgb}{0.21,0.49,0.74}
\usepackage[pagebackref,breaklinks,colorlinks,allcolors=cvprblue]{hyperref}

\def\paperID{5}
\def\confName{CVPR}
\def\confYear{2026}

\title{Controllable Image Generation with Composed Parallel Token Prediction}

\author{
    Jamie Stirling\\
Durham University\\
United Kingdom\\
{\tt\small jamie.s.stirling@durham.ac.uk}
\and
Noura Al-Moubayed\\
Durham University\\
United Kingdom\\
{\tt\small noura.al-mouyed@durham.ac.uk}
\and
Chris G. Willcocks\\
Durham University\\
United Kingdom\\
{\tt\small christopher.g.willcocks@durham.ac.uk}
\and
Hubert P. H. Shum\\
Durham University\\
United Kingdom\\
{\tt\small hubert.shum@durham.ac.uk}
}

\begin{document}
\maketitle
\input{cvpr-2026/sec/0_abstract}    
\input{cvpr-2026/sec/1_intro}
\input{cvpr-2026/sec/2_related}

\input{cvpr-2026/sec/3_methods}
\input{cvpr-2026/sec/4_experiments}
\input{cvpr-2026/sec/5_discussion}

\input{cvpr-2026/sec/6_conclusion}

{
    \small
    \bibliographystyle{unsrtnat}
    \bibliography{cvpr-2026/main}
}

% \input{cvpr-2026/sec/X_suppl}

\end{document}

%% file: cvpr-2026/sec/0_abstract.tex
\begin{abstract}
  Conditional discrete generative models struggle to faithfully compose multiple input conditions. To address this, we derive a theoretically-grounded formulation for composing discrete probabilistic generative processes, with masked generation (absorbing diffusion) as a special case. Our formulation enables precise specification of novel combinations and numbers of input conditions that lie outside the training data, with concept weighting enabling emphasis or negation of individual conditions. In synergy with the richly compositional learned vocabulary of VQ-VAE and VQ-GAN, our method attains a $63.4\%$ relative reduction in error rate compared to the previous state-of-the-art, averaged across 3 datasets (positional CLEVR, relational CLEVR and FFHQ), simultaneously obtaining an average absolute FID improvement of $-9.58$. Meanwhile, our method offers a $2.3\times$ to $12\times$ real-time speed-up over comparable methods, and is readily applied to an open pre-trained discrete text-to-image model for fine-grained control of text-to-image generation.
\end{abstract}

%% file: cvpr-2026/sec/1_intro.tex
\section{Introduction}

% Introduce the main and relevant concepts and maybe a bit of history of the field?
Conditional image generation models struggle to faithfully satisfy multiple input conditions simultaneously, especially for novel combinations outside the training data \cite{NEURIPS2023_f8ad010c} (compositional generalisation \cite{lin2023survey}).

\begin{figure}[t]
    \centering
    \begin{subfigure}[t]{0.155\textwidth}
        \includegraphics[trim=0.25cm 0cm 0.25cm 0cm, clip, width=\textwidth]{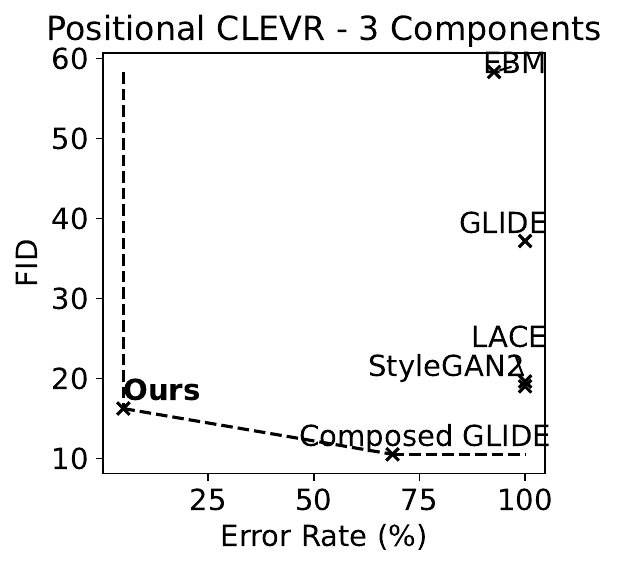}
        \raggedright
    \end{subfigure}
    \begin{subfigure}[t]{0.155\textwidth}
        \includegraphics[trim=0.25cm 0cm 0.25cm 0cm, clip, width=\textwidth]{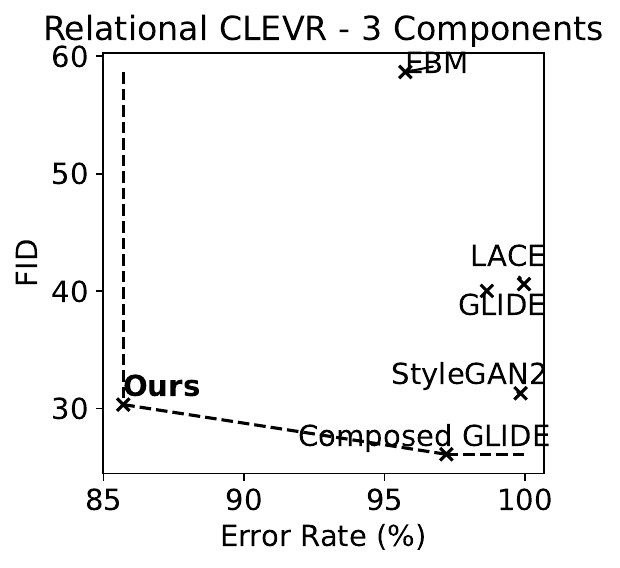}
    \end{subfigure}
    \begin{subfigure}[t]{0.155\textwidth}
        \includegraphics[trim=0.25cm 0cm 0.25cm 0cm, clip, width=\textwidth]{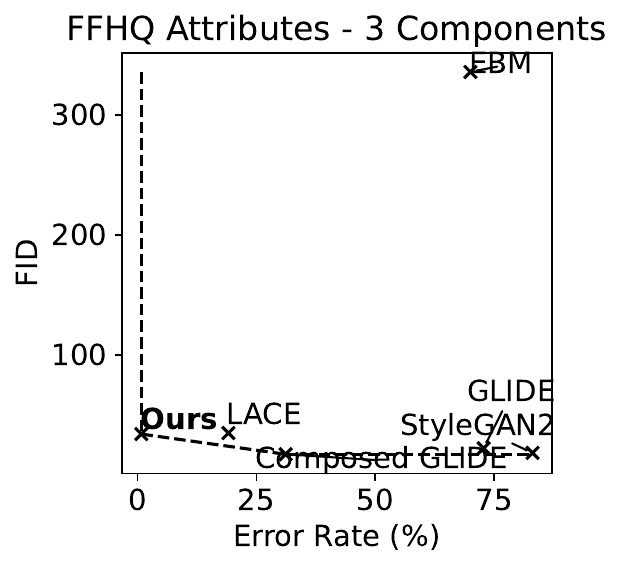}
    \end{subfigure}
    \caption{Scatter plots of compositional generation error vs FID on 3 datasets (3 input components). Our method lies on the Pareto front of all results (see Appendix for full scatter plots) while achieving the lowest or joint lowest error among the baselines.}
    \label{fig:intro_scatter}
    \vspace{-0em}
\end{figure}

\begin{figure}[t]
\centering
\begin{subfigure}[t]{0.325\linewidth}
\includegraphics[width=\textwidth]{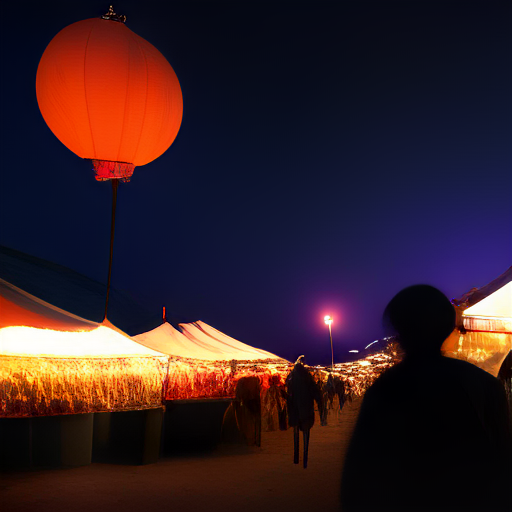}
\end{subfigure}
\hfill
\begin{subfigure}[t]{0.325\linewidth}
\includegraphics[width=\textwidth]{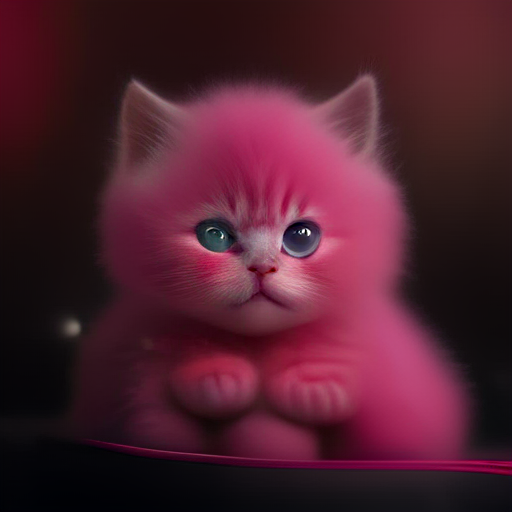}
\end{subfigure}
\hfill
\begin{subfigure}[t]{0.325\linewidth}
\includegraphics[width=\textwidth]{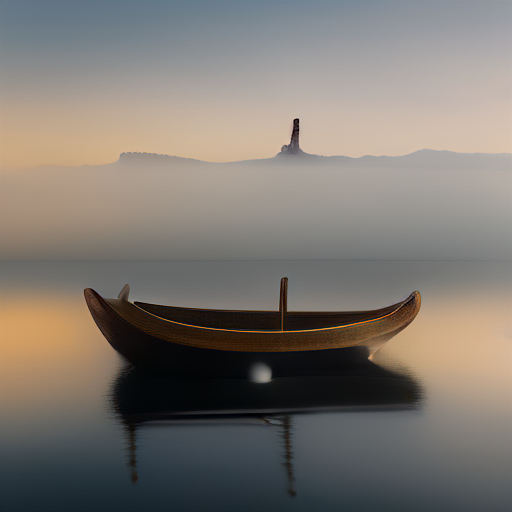}
\end{subfigure}
\begin{subfigure}[t]{0.325\linewidth}
\includegraphics[width=\textwidth]{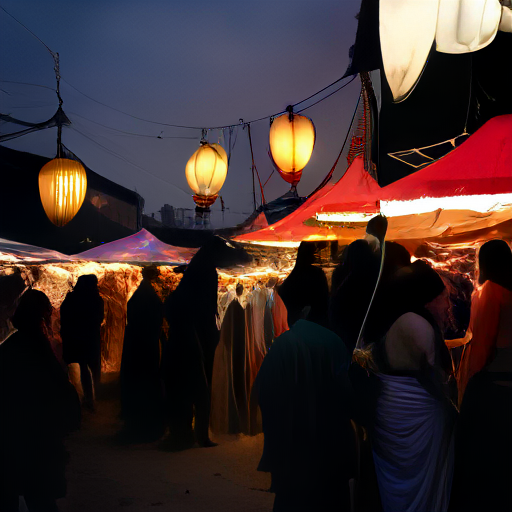}
\caption*{\scriptsize "Vibrant, bustling outdoor market with colourful stalls" \textbf{AND} "Vendors and shoppers from various cultures interacting" \textbf{AND} "Hanging lanterns illuminating the scene"}
\end{subfigure}
\hfill
\begin{subfigure}[t]{0.325\linewidth}
\includegraphics[width=\textwidth]{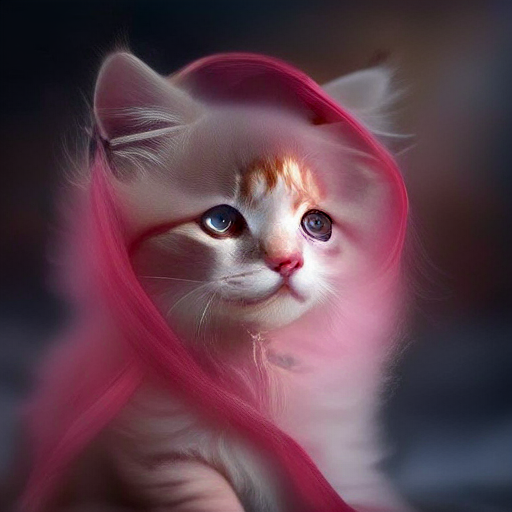}
\caption*{\scriptsize "Fluffy kitten with big eyes and pink nose" \textbf{AND} "Kitten tangled in ball of red yarn" \textbf{AND} "Soft lighting casting gentle shadows"}
\end{subfigure}
\hfill
\begin{subfigure}[t]{0.325\linewidth}
\includegraphics[width=\textwidth]{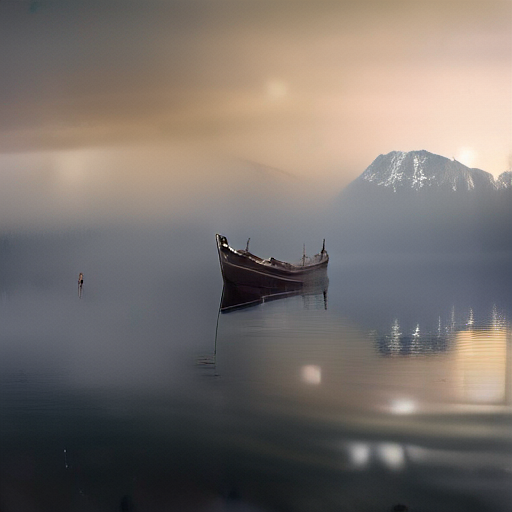}
\caption*{\scriptsize "Rustic, wooden rowboat floating on misty lake" \textbf{AND} "Snow-capped mountains in the distance, reflected in water" \textbf{AND} "Morning light filtering through mist, golden glow"}
\end{subfigure}
  \caption{Compositional text-to-image results with captions (zooming recommended). \textit{Top:} single-prompt baseline. \textit{Bottom:} composed multi-prompt (ours). Our method enables the composition of multiple conditions, conferring an advantage over the single-prompt baseline. }
\label{fig:qual_results_with_captions}
\end{figure}

\begin{figure*}[t]
    \centering
    \begin{subfigure}[t]{0.16\textwidth}
        \includegraphics[width=\textwidth]{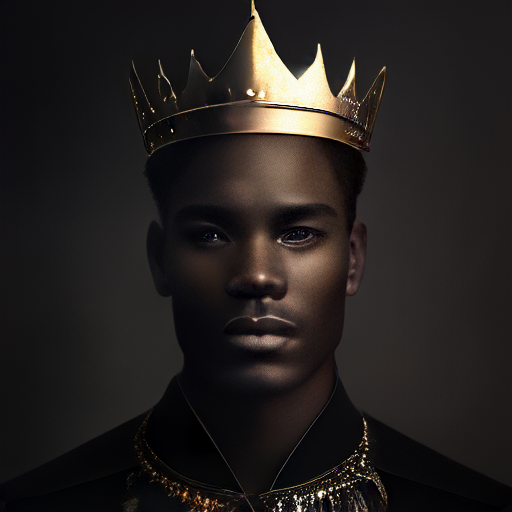}
        \raggedright
        \caption*{\scriptsize [*] "a king \textbf{not} wearing a crown, portrait"}
    \end{subfigure}
    \hfill
    \begin{subfigure}[t]{0.16\textwidth}
        \includegraphics[width=\textwidth]{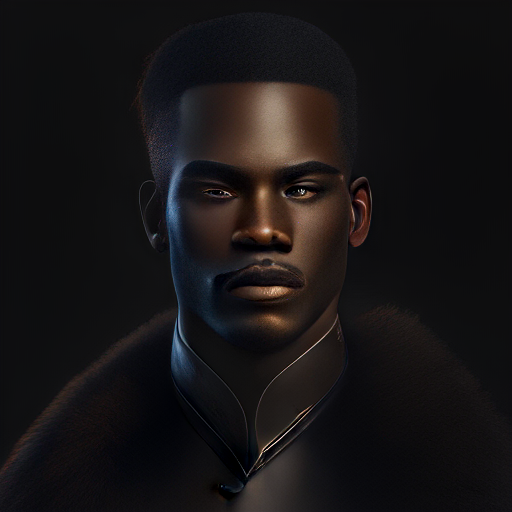}
        \caption*{\scriptsize "a king, portrait" \textbf{AND} (\textbf{NOT} "wearing a crown")}
        \end{subfigure}
    \hfill
    \begin{subfigure}[t]{0.16\textwidth}
        \includegraphics[width=\textwidth]{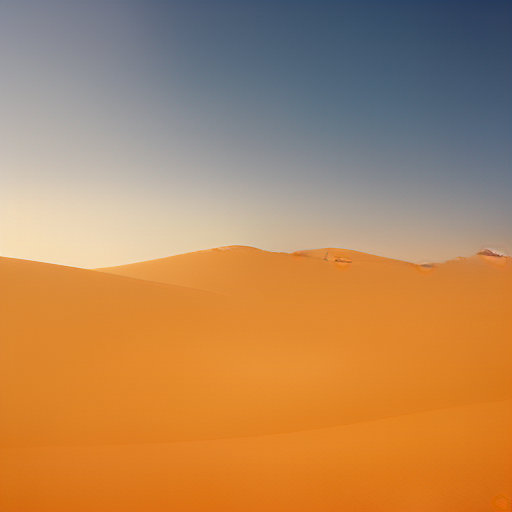}
        \caption*{\scriptsize [*] "a sunny desert with \textbf{no} sand dunes, landscape" }
    \end{subfigure}
    \hfill
    \begin{subfigure}[t]{0.16\textwidth}
        \includegraphics[width=\textwidth]{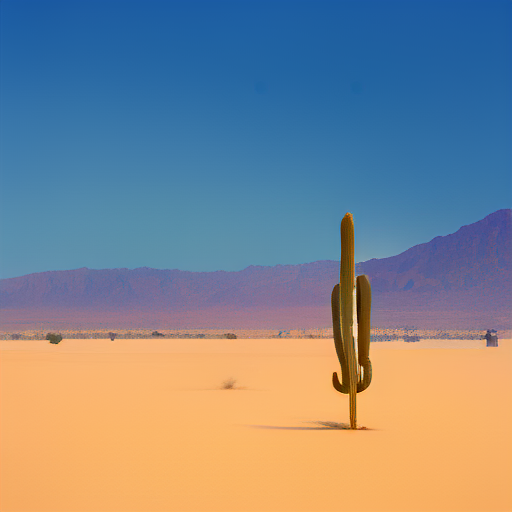}
        
        \caption*{\scriptsize "a sunny desert, landscape" \textbf{AND} (\textbf{NOT} "sand dunes")}
    \end{subfigure}
    \hfill
    \begin{subfigure}[t]{0.16\textwidth}
        \includegraphics[width=\textwidth]{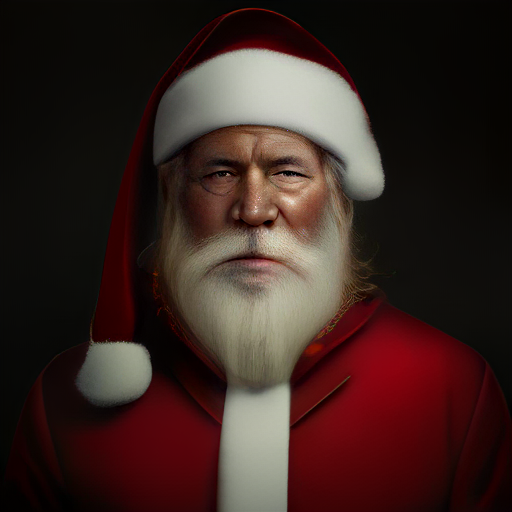}
        
        \caption*{\scriptsize [*] "santa claus \textbf{not} wearing red, portrait"}
    \end{subfigure}
    \hfill
    \begin{subfigure}[t]{0.16\textwidth}
        \includegraphics[width=\textwidth]{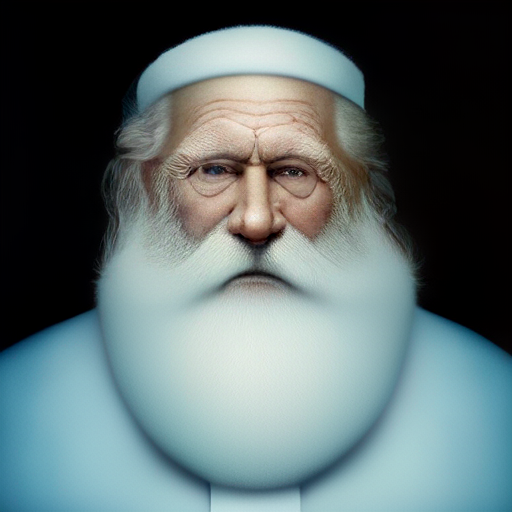}
        \caption*{\scriptsize "santa claus, portrait" \textbf{AND} (\textbf{NOT} "wearing red")}
    \end{subfigure}

  \caption{ Concept negation with text-to-image (left baseline, right ours): Our method allows more precise control over the outputs of an existing pre-trained model (aMUSEd \cite{patil2024amused}).  The baseline handles poorly the negation/removal of characteristics which commonly co-occur with the subject of the image (e.g. a king with \textit{no} crown). Each image is selected from three runs as the most representative of the prompt (ours and baseline).}
    \label{fig:concept_negation}
    \vspace{-0em}
\end{figure*}
In the area of continuous iterative image generation approaches, earlier works \cite{nie2021controllable,du2020compositional,liu2022compositional} have proposed methods for improving controllability of image generation via \textit{composition} of energy-based models and diffusion models. These models improve condition-output alignment, significantly exceeding non-composed baselines in terms of accuracy and image quality \cite{liu2022compositional}. However, such approaches do not extend to \textit{discrete} sample spaces, which offer a number of trade-offs and outright improvements over their continuous counterparts, including data-efficiency and temperature control\cite{van2017neural,esser2020taming,bond2021unleashing,chang2023muse}.

Discrete image generation methods, including autoregressive \cite{esser2020taming} and masked sampling \cite{bond2021unleashing,chang2022maskgit,chang2023muse,patil2024amused}, offer advantages in generation speed, image quality and diversity over continuous analogues \cite{du2019implicit,ho2020denoising,lipman2022flow}. Until now, these advantages have remained mutually exclusive from those of compositional generation, which have thus far been limited to \textit{continuous} approaches only \cite{du2020compositional,liu2022compositional}.

We address this gap by deriving a theoretically-grounded framework for composing discrete generative model outputs, enabling faithful multi-condition generation. Applying this framework to conditional parallel token prediction (absorbing diffusion), we leverage the expressive  and richly compositional ``visual language'' emergent in discrete representation methods like VQ-VAE and VQ-GAN. Concept weighting provides an additional degree of controllability, enabling emphasis or negation of individual conditions.

\begin{figure*}[t]
    \centering
    \includegraphics[width=0.7\linewidth]{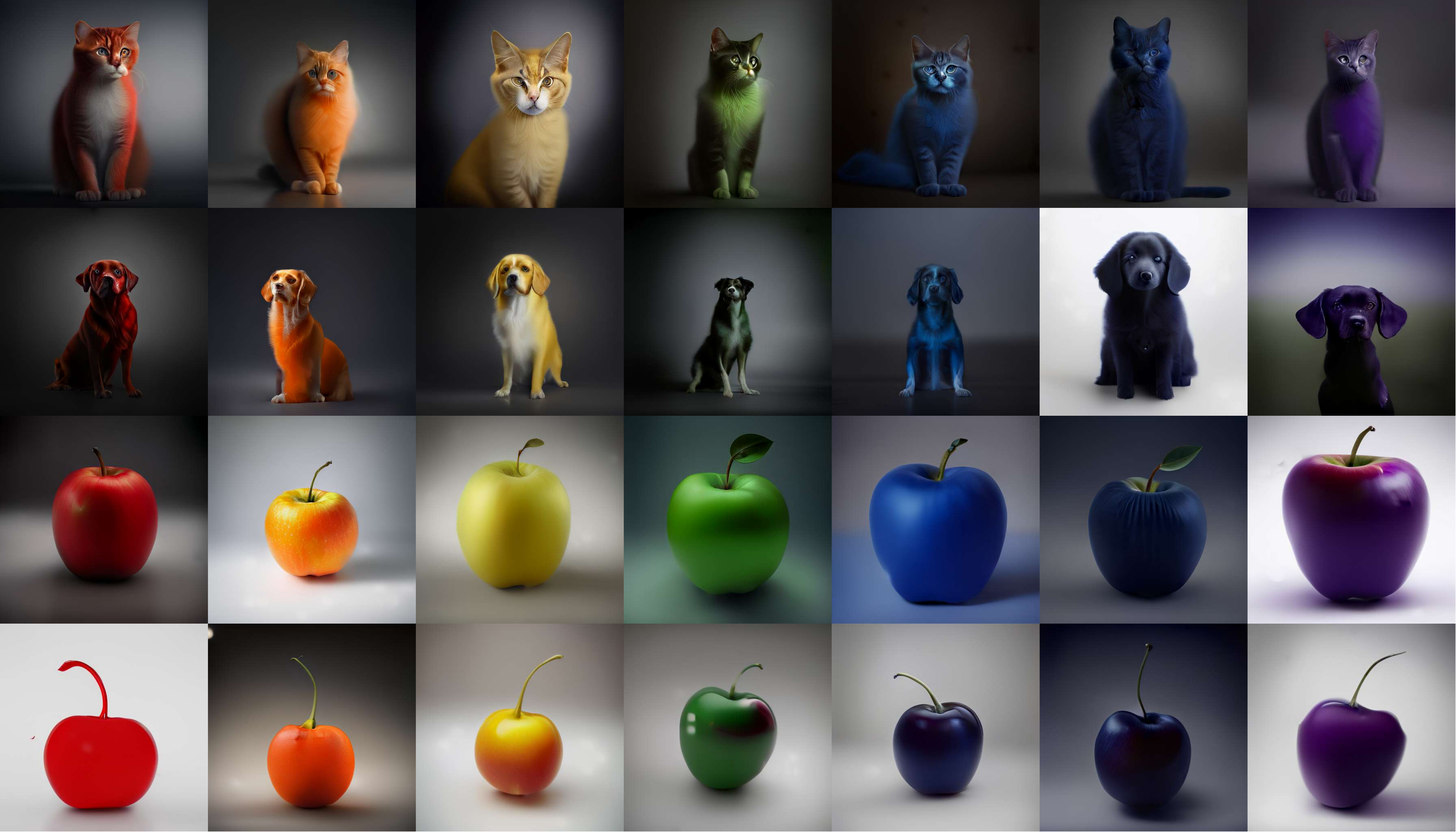}
    \caption{Conceptual product space: Example of composing two concept spaces using our framework: \{"a cat","a dog","an apple","a cherry"\}$\times_\mathrm{\scaleto{\boldsymbol{AND}}{4px}}$\{"a red object","an orange object","a yellow object","a green object","a blue object","an indigo object","a violet object"\}. }
    \label{fig:concept-grid-2-4}
    \vspace{-0em}
\end{figure*}

We train and evaluate our approach on positional CLEVR \cite{johnson2017clevr, liu2022compositional}, relational CLEVR \cite{johnson2017clevr, liu2022compositional} and FFHQ \cite{karras2019style}. Our method achieves state-of-the-art error rates across all 9 quantitative settings (3 datasets with $1$, $2$, or $3$ input components), as shown in Figure \ref{fig:intro_scatter}, equating to an average relative error rate reduction of 63.4\%. We outperform the next-best method (ranked by error rate) in terms of FID for 7 out of 9 settings, meanwhile maintaining the significant speed advantages of parallel token prediction, offering a $2.3\times$ to $12\times$ speed-up in real time over comparable continuous approaches. We further show that our method can be applied to an open pre-trained text-to-image parallel token model (aMUSEd \cite{patil2024amused}) with appealing visual results (Figures \ref{fig:qual_results_with_captions, fig:concept-grid-2-4}).

%\iffalse
The contributions of this research are summarised as:
\begin{itemize}
\item Extending the product-of-experts principle, we derive a theoretically-grounded formulation for composing conditional distributions in the discrete sequential generation setting. This result is generalised to arbitrary discrete iterative processes, encompassing both masked and autoregressive sampling.
\item We adapt the formulation to conditional parallel token prediction (absorbing diffusion \cite{bond2021unleashing}), enabling efficient, high-quality, and controllable image synthesis in the discrete latent space of VQ-VAE and VQ-GAN.
\item We demonstrate empirically that our method achieves state-of-the-art accuracy and competitive FID scores across 3 datasets, while offering a $2.3\times$ to $12\times$ speed-up in real time, over comparable continuous methods.
\item We release source code of our method under the MIT license.
\end{itemize}

%% file: cvpr-2026/sec/2_related.tex
\section{Related Work}
\label{sec:related}
We examine the literature from the perspectives of \textit{compositional methods} and \textit{discrete generative methods}, with our work lying in the intersection of the two.

\textbf{Continuous compositional generation: }
Earlier works proposed methods for conditioning continuous generative models \cite{du2020compositional,nie2021controllable,liu2022compositional} based on the conjunction and negation of input attributes, drawing on formal analogues to product-of-experts \cite{Hinton1999ProductsExperts}. To the best of our knowledge, no analogous method exists for composing flow matching models \cite{lipman2022flow}. \cite{liu2022compositional} introduces an approach that enhances the capabilities of text-conditioned diffusion models in generating complex and photo-realistic images based on textual descriptions.

Many compositional approaches apply a weighting factor to each component, which resembles the mathematical form of \textbf{classifier-free guidance (CFG)} which is readily applied to both continuous and discrete image generation \cite{ho2022classifier, patil2024amused}.

Applications-wise, \cite{wu2024compositional} proposes applying composed diffusion to a multi-constraint design task, which is shown to generalise to designs which are more complex than those in the training data (compositional generalization). Similarly, \cite{zhang2025energymogen} composes multiple energy terms to constrain the generation of motion from text. These methods demonstrate the practical utility and contemporary relevance of compositional approaches.

Beyond the literature, our work diverges significantly from existing work with continuous methods, in providing the first (to the best of our knowledge) formulation for composing arbitrary \textit{discrete} generative processes.

\textbf{Composition in sequential tasks: }
Earlier work in reinforcement learning \cite{NEURIPS2019_95192c98} has sought to use ideas relevant to our own for composing \textit{policies} for the purposes of compositional generalisation in multi-timestep environments. The main idea shared with our work is that of \textit{multiplying} constituent ``primitives'' (as opposed to \textit{additive} composition, as in mixture-of-experts \cite{masoudnia2014mixture}). This shares some superficial similarities with our work on compositional discrete image generation, where multiple attributes are required to be expressed in the same output image; however, it differs significantly in both the application and the formalism.

\textbf{Discrete representation learning and sampling: }
Discrete representation learning has emerged as the discrete counterpart to continuous VAE approaches \cite{van2017neural,razavi2019generating}. Discrete representation learning is based on the concept of vector-quantization (VQ) \cite{gray1984vector}, whereby features from a continuous vector space are mapped to an element of a finite set of learned codebook vectors. This VQ family of models includes VQ-VAE \cite{van2017neural} and VQ-GAN \cite{esser2020taming}. VQ-family approaches require a secondary prior model to be trained to sample from the discrete latent space, which can be computationally expensive at both train- and inference- time \cite{van2017neural,esser2020taming,bond2021unleashing}. An alternative approach is to predict multiple tokens in parallel transformer encoder \cite{bond2021unleashing,chang2023muse,patil2024amused} (akin to masked language modelling \cite{devlin2018bert}), which introduces a controllable trade-off between sample speed and generation quality. The per-image generation time is still linear in the size ($W\times H$) of the image, albeit with a smaller constant than autoregressive models \cite{bond2021unleashing}. Masked prediction can be further extended to unaligned conditional 2D-3D image synthesis \cite{corona2023unaligned}.

Beyond autoregressive and masked generation approaches, there exists a handful of alternative discrete generative methods. \textit{Discrete flow matching} \cite{gat2024discrete} and bitwise autoregressive modelling \cite{han2025infinity} are notable recent examples.

%% file: cvpr-2026/sec/3_methods.tex
\begin{figure*}[t]
\centering
    \includegraphics[width=0.9\textwidth]{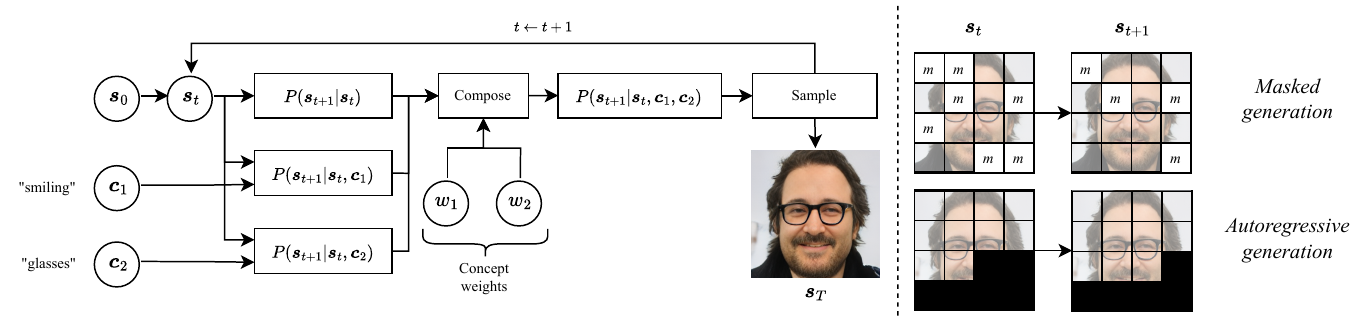}
\caption{Overview of our approach. \textit{Left}: We start with a generic ``empty'' state $\boldsymbol{s}_0$ or the preceding state $\boldsymbol{s}_t$ and a set of input concepts $\boldsymbol{\boldsymbol{c}_1,\boldsymbol{c}_2,...}$. These are used to compute the unconditional distribution over $\boldsymbol{s}_{t+1}$ and the conditional distributions given each of $\boldsymbol{\boldsymbol{c}_1,\boldsymbol{c}_2,...}$. These are \textit{composed}, optionally incorporating \textit{concept weights}, to produce an accurate estimate of the distribution conditioned on \textit{all} inputs. The process is repeated until a ``full'' state $\boldsymbol{s}_T$ is obtained, e.g. a grid of VQ-VAE latents. \textit{Right}: Our general formulation encompasses any \textit{additive} discrete generative process, i.e. where the representation of $\boldsymbol{s}_t$ is fully determined by $\boldsymbol{s}_{t+1}$ (masked and autoregressive generation shown as examples).}
    \label{fig:overview}
    \vspace{-0em}
\end{figure*}

\section{Method}

\label{sec:methods}
In this section, we present our method for composing generative models for controllable sampling from discrete representation spaces of images. 

The overview of our approach is shown in Figure \ref{fig:overview}. We first derive a novel formulation that directly informs our method of composing conditional distributions over discrete spaces. Next, we show how this extends generally to all discrete sequential generation tasks, in which categorical variables are sampled iteratively to produce a complete sample (e.g. via autoregressive or masked models). We show how this result can be specifically adapted for conditional parallel token prediction \cite{bond2021unleashing,chang2023muse} to achieve high-quality and accurate controllable image synthesis. This is further enhanced by concept weighting, which allows the relative importance of input conditions to be increased, decreased or negated to the desired effect. We note that similar results can be obtained for other types of generative models (provided they are iterative and discrete, see Appendix for examples). This section lays the groundwork for our later experiments with compositional sampling from the latent space of VQ-VAE and VQ-GAN.

% Same as many previous works; details in supplementary
% In some cases using pre-trained (with this method or other methods)
\subsection{Composing Conditional Categorical Distributions}
% state our hypothesis/modelling assumption
Our framework is derived from the simplifying assumption employed by the product of experts \cite{Hinton1999ProductsExperts}, in which the input conditions $\boldsymbol{c}_1,...,\boldsymbol{c}_n$ are independent of each other conditional on the output variable $\boldsymbol{x}$, i.e. $P(\boldsymbol{c}_1,\boldsymbol{c}_2|\boldsymbol{x})\propto P(\boldsymbol{c}_1|\boldsymbol{x})P(\boldsymbol{c}_2|\boldsymbol{x})$ for any pair of distinct attributes $(\boldsymbol{c}_1,\boldsymbol{c}_2)$. A consequence of this is that the probability of an outcome $\boldsymbol{x}$ given two or more conditions $\boldsymbol{c}_1,\boldsymbol{c}_2,...$ is proportional to the product of the probabilities of each condition given $\boldsymbol{x}$. Intuitively, taking the product of different categorical distributions in this way is analogous to taking the intersection of the sample spaces of two or more conditional distributions, thus (ideally) resulting in samples which embody all of the specified conditions. We discuss the benefits and limitations of this in Section \ref{sec:discussion}. Following previous work with composition of continuous models \cite{xu2022compositional,liu2022compositional}, we factorize the distribution of a $k$-way categorical variable $\boldsymbol{x}$ conditioned on $n$ variables as
\begin{equation}
    P(\boldsymbol{x}|\boldsymbol{c}_1,...,\boldsymbol{c}_n) \propto P(\boldsymbol{x})\prod_{i=1}^nP(\boldsymbol{c}_i|\boldsymbol{x}).
    \label{eq:decomp}
\end{equation}

Applying Bayes' theorem \cite{joyce2003bayes}, this can be re-written as
\begin{equation}
\begin{aligned}
    P(\boldsymbol{x}|\boldsymbol{c}_1,...,\boldsymbol{c}_n) &\propto P(\boldsymbol{x})\prod_{i=1}^n\frac{P(\boldsymbol{x}|\boldsymbol{c}_i)P(\boldsymbol{c}_i)}{P(\boldsymbol{x})},
    \label{eq:bayes}
    \\
    &\propto P(\boldsymbol{x})\prod_{i=1}^n\frac{P(\boldsymbol{x}|\boldsymbol{c}_i)}{P(\boldsymbol{x})}.
\end{aligned}
\end{equation}

We are able to eliminate the $P(\boldsymbol{c}_i)$ term in (\ref{eq:bayes}) as a consequence of normalising the values of $P(\boldsymbol{x}=x_i|...)$ for all $x_i$, such that they sum to $1$. Please refer to Appendix for the full derivation.

\subsection{Composition for Sequential Generative Tasks}

So far, we have shown how our approach applies to conditional generation with a single categorical output $\boldsymbol{x}$. In practice, many generative tasks involve sampling multiple categorical variables (tokens or latent codes) over a number of time steps \cite{Brown2020LanguageLearners, Vaswani2017AttentionNeed} where the sampling of a new state $\boldsymbol{s}_{t+1}$ at each successive step $t$ is conditioned on the previous state (in addition to the specified conditions $\boldsymbol{c}_1,...,\boldsymbol{c}_n$). Formulating this alongside the result in \eqref{eq:bayes} gives the following general expression for composing discrete sequential generation tasks (see Appendix for further explanation):
\begin{equation}
    P(\boldsymbol{s}_{t+1}|\boldsymbol{s}_{t},\boldsymbol{c}_1,...,\boldsymbol{c}_n)\propto P(\boldsymbol{s}_{t+1}|\boldsymbol{s}_{t})\prod_{i=1}^n\frac{P(\boldsymbol{s}_{t+1}|\boldsymbol{s}_{t},\boldsymbol{c}_i)}{P(\boldsymbol{s}_{t+1}|\boldsymbol{s}_{t})}.
    \label{eq:sequential}
\end{equation}

We observe that this applies generally to any generative process in which each successive step is conditioned on the output of previous steps, including autoregressive language modelling \cite{Vaswani2017AttentionNeed}, masked language modelling \cite{Devlin2018BERT:Understanding}, as well as autoregressive \cite{esser2020taming} and non-autoregressive \cite{bond2021unleashing} approaches for image generation. In the remainder of this paper, we maintain a particular focus on conditional parallel (masked) token prediction \cite{bond2021unleashing}, which we use for composed sampling from the latent space of VQ-VAE \cite{van2017neural} and VQ-GAN \cite{esser2020taming} for high-fidelity image synthesis.

A key practical consideration concerning the result in \eqref{eq:sequential} is that estimates must be obtained for each conditional probability distribution $P(\boldsymbol{s}_{t+1}|\boldsymbol{s}_{t},\boldsymbol{c}_i)$ in addition to $P(\boldsymbol{s}_{t+1}|\boldsymbol{s}_{t})$. In each of our experiments (Section \ref{sec:experiments}), we ensure that, during training, conditional information is zero-masked with a set probability ($0.1$) per sample, thus allowing us to obtain $P(\boldsymbol{s}_{t+1}|\boldsymbol{s}_{t})$ at inference time by supplying zeros in place of the condition encoding.

\subsection{Composed Parallel Token Prediction}
% Then extend to the step-by-step generation process with masked encoder
Parallel token prediction \cite{bond2021unleashing,chang2022maskgit,chang2023muse} is a non-autoregressive alternative to next-token prediction \cite{esser2020taming} for generative sampling from a discrete latent space. This allows a direct trade-off between sampling speed and image generation quality \cite{bond2021unleashing} by controlling the rate at which tokens are sampled. 

Parallel token prediction can be thought of as the gradual un-masking of a collection of discrete latent codes $\boldsymbol{z}_0$ given the partial reconstruction from a previous time step (as well as additional conditioning information). This corresponds directly to the next-state prediction formulation in \eqref{eq:sequential}, but with the time labels $t$ reversed in order to reflect the ``reverse process'' which characterizes diffusion models (both continuous \cite{dhariwal2021diffusion} and discrete \cite{bond2021unleashing}):
\begin{equation}
    P(\boldsymbol{z}_{t-1}|\boldsymbol{z}_{t},\boldsymbol{c}_1,...,\boldsymbol{c}_n)\propto P(\boldsymbol{z}_{t-1}|\boldsymbol{z}_{t})\prod_{i=1}^n\frac{P(\boldsymbol{z}_{t-1}|\boldsymbol{z}_{t},\boldsymbol{c}_i)}{P(\boldsymbol{z}_{t-1}|\boldsymbol{z}_{t})}.
    \label{eq:parallel-token}
\end{equation}
\begin{figure*}[t]
    \centering
    \includegraphics[width=0.8\linewidth]{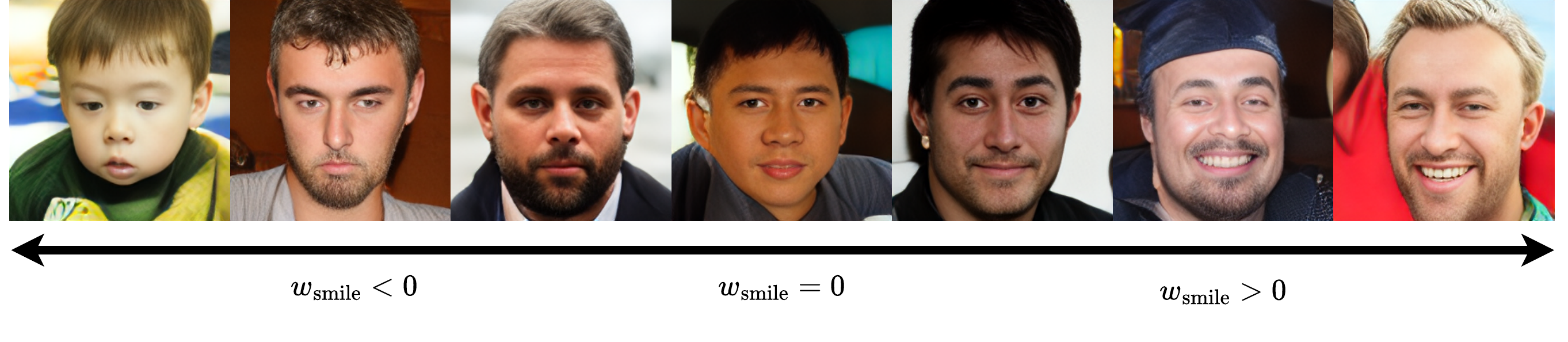}
    \caption{Effect of varying the $w_\textrm{smile}$ concept weight from $-3.0$ to $3.0$ while keeping $w_\textrm{male}=w_\textrm{no\_glasses}=3.0$. }
    \label{fig:vary_w_smile}
\end{figure*}

In \eqref{eq:parallel-token}, $\boldsymbol{z}_t$ is an intermediate, partially unmasked representation at each time step, and $\boldsymbol{z}_{t-1}$ represents the distribution over image representations with fewer masked tokens. In practise, following earlier work with parallel token prediction, the model is trained to directly predict the fully unmasked representation $\boldsymbol{z}_0$  (as opposed to intermediate states) so as to maximise training stability \cite{bond2021unleashing}. At inference time, we compute the composed unmasking probabilities as:

\begin{equation}
    P(\boldsymbol{z}_{0}|\boldsymbol{z}_{t},\boldsymbol{c}_1,...,\boldsymbol{c}_n)\propto P(\boldsymbol{z}_{0}|\boldsymbol{z}_{t})\prod_{i=1}^n\frac{P(\boldsymbol{z}_{0}|\boldsymbol{z}_{t},\boldsymbol{c}_i)}{P(\boldsymbol{z}_{0}|\boldsymbol{z}_{t})}.
    \label{eq:parallel-token-inference}
\end{equation}
Image representations are then unmasked one or more tokens at a time, corresponding to a trade-off between sample speed (more tokens per iteration) and image generation quality (fewer tokens per iteration) \cite{bond2021unleashing}.

\subsection{Concept Weighting to Improve Controllability}
We introduce an additional set of hyperparameters $w_1,...w_n$ to specify the relative importance of each condition $\boldsymbol{c}_1,...,\boldsymbol{c}_n$ respectively, motivated by similar research employing diffusion models for image generation \cite{liu2022compositional}. Restating \eqref{eq:sequential} in terms of log-probabilities and introducing these weighting terms gives:
\begin{equation}
\begin{split}
    &\log P(\boldsymbol{s}_{t+1}|\boldsymbol{s}_{t},\boldsymbol{c}_1,...,\boldsymbol{c}_n)\\=& \log P(\boldsymbol{s}_{t+1}|\boldsymbol{s}_{t})\\&+ \sum_{i=1}^n w_i\left[\log P(\boldsymbol{s}_{t+1}|\boldsymbol{s}_{t},\boldsymbol{c}_i) - \log P(\boldsymbol{s}_{t+1}|\boldsymbol{s}_{t})\right].
    \label{eq:weighted}
\end{split}
\end{equation}

This expression \eqref{eq:weighted} can be readily reformulated into the corresponding form for parallel token prediction \eqref{eq:parallel-token}. A key observation here is that setting a concept weight $w_i$ to \textit{negative} (e.g. $-1$) has the intuitive effect of \textit{negating} the corresponding condition $\boldsymbol{c}_i$ by excluding image representations which correspond to $\boldsymbol{c}_i$ from the sample space. Altogether, the prompt-weighting approach provides an additional degree of controllability over model outputs by enabling conditions to be emphasized ($w_i>1$), de-emphasized ($w_i<1$) or even negated ($w_i<0$) as required. We demonstrate the practical utility of this feature in our qualitative experiments (Section \ref{sec:experiments}). We do not include \textit{disjunction} in our evaluation for reasons explained in Appendix.

In practice, we use our compositional framework to sample from the discrete latent space of VQ-VAE \cite{van2017neural} and VQ-GAN \cite{esser2020taming}, which are powerful and practical approaches for encoding images and other high-dimensional modalities as collections of discrete latent codes (visual tokens) while producing high-fidelity reconstructions and generated samples. 

\subsection{Discrete Encoding and Decoding of Images}
In order to compose categorical distributions for generating images, we must also define an invertible mapping between RGB images and discrete latent representations. We utilise convolutional down-sampling and up-sampling (autoencoder) to map between RGB image space and latent embedding space. Following the original VQ-VAE \cite{van2017neural} formulation, we employ nearest-neighbour vector quantization, in which encoder outputs are mapped to their nearest neighbour in a learned codebook. Specifically, for each encoder output vector $z_e$, the corresponding quantized vector is computed as the nearest codebook entry $e_c$, where
\begin{equation}
    c=\arg\min_j||z_e-e_j||_2
\end{equation}
and $e_0,e_1...e_{K-1}$ are entries in a learned vector codebook of length $K$.

Since the quantization step is non-differentiable, it is necessary to estimate the gradients during backpropagation. For this purpose, we apply straight-through gradient estimation \cite{bengio2013estimating}, whereby during backpropagation, the gradients are copied directly from the decoder input $z_c$ to the encoder output $z_e$. We adopt this vector quantization approach for all experiments, including the embedding and commitment loss terms from the original VQ-VAE formulation \cite{van2017neural}.

%% file: cvpr-2026/sec/4_experiments.tex
\begin{figure*}[t]
    \centering
    \begin{subfigure}[t]{0.13\textwidth}
        \includegraphics[width=\textwidth]{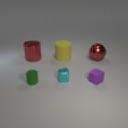}
        \raggedright
        \caption*{\scriptsize (0.25, 0.6), (0.5, 0.6),\\(0.75, 0.6), (0.25, 0.4),\\(0.5, 0.4), (0.75, 0.4)}
    \end{subfigure}
    \hfill
    \begin{subfigure}[t]{0.13\textwidth}
        \includegraphics[width=\textwidth]{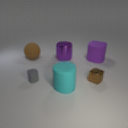}
        \caption*{\scriptsize (0.25, 0.6), (0.5, 0.6),\\(0.75, 0.6), (0.25, 0.4),\\(0.5, 0.4), (0.75, 0.4)}
        \end{subfigure}
    \hfill
    \begin{subfigure}[t]{0.13\textwidth}
        \includegraphics[width=\textwidth]{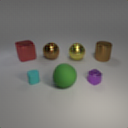}
        \caption*{\scriptsize (0.2, 0.6), (0.4, 0.6),\\(0.6, 0.6), (0.8, 0.6),\\(0.25, 0.4), (0.5, 0.4),\\(0.75,0.4)}
    \end{subfigure}
    \hfill
    \begin{subfigure}[t]{0.13\textwidth}
        \includegraphics[width=\textwidth]{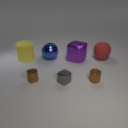}
        \caption*{\scriptsize (0.2, 0.6), (0.4, 0.6),\\(0.6, 0.6), (0.8, 0.6),\\(0.25, 0.4), (0.5, 0.4),\\(0.75,0.4)}
    \end{subfigure}
    \hfill
    \begin{subfigure}[t]{0.13\textwidth}
        \includegraphics[width=\textwidth]{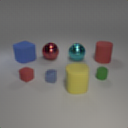}
        
        \caption*{\scriptsize (0.2, 0.6), (0.4, 0.6),\\(0.6, 0.6), (0.8, 0.6),\\(0.2, 0.4), (0.4, 0.4),\\(0.6, 0.4), (0.8, 0.4)}
    \end{subfigure}
    \hfill
    \begin{subfigure}[t]{0.13\textwidth}
        \includegraphics[width=\textwidth]{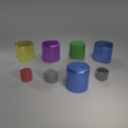}
        \caption*{\scriptsize (0.2, 0.6), (0.4, 0.6),\\(0.6, 0.6), (0.8, 0.6),\\(0.2, 0.4), (0.4, 0.4),\\(0.6, 0.4), (0.8, 0.4)}
    \end{subfigure}

  \caption{ Compositional out-of-distribution generation: Positional CLEVR training images contain no more than 5 objects per image, but our compositional method allows 6 or more objects to appear in the same image via compositional sampling. }
    \label{fig:out_of_distribution}
    \vspace{-1.2em}
\end{figure*}
\section{Experiments}
\label{sec:experiments}

\subsection{Datasets}
Following earlier work \cite{liu2022compositional} in evaluating compositional generalisation for image generation, we employ three datasets for training and evaluation: Positional CLEVR \cite{johnson2017clevr,liu2022compositional}, Relational CLEVR \cite{johnson2017clevr,liu2022compositional} and FFHQ \cite{karras2019style} (full description of datasets in Appendix). These three datasets are chosen to represent a range of unique and challenging compositional tasks (conditioned on object position, object relations, and facial attributes respectively). For each of the three datasets, we train a VQ-VAE or VQ-GAN model to enable encoding and decoding between the image space and the discrete latent representation space, in addition to a conditional parallel token prediction model (encoder-only transformer) which learns to unmask discrete latent representations, optionally conditioned on an encoded input annotation.

\subsection{Model Training}
We train a VQ-GAN model to reconstruct FFHQ samples at $256\times256$ resolution, as well as VQ-VAE models for each of CLEVR and Relational CLEVR at $128\times128$ resolution. These choices of resolution follow earlier work in compositional generation with these 3 datasets \cite{liu2022compositional}. We find in practice that VQ-VAE (without the adversarial loss) is sufficient for high-fidelity reconstruction of the two CLEVR datasets due to the smaller resolution and visual simplicity, while VQ-GAN is required for realistic reconstructions of FFHQ. Unlike \cite{liu2022compositional}, our choice of training regime produces FFHQ images directly at $256\times256$, so a post-upsampling step is not required during evaluation. We train with a perceptual loss \cite{zhang2018unreasonable} in addition to the MSE loss for all datasets (and the learned adversarial loss for FFHQ). Full details of model training are in Appendix.

\subsection{Quantitative Evaluation of Compositional Generation}
For each dataset, following \cite{liu2022compositional} we evaluate (compositionally) generated image samples according to both FID (Fréchet Inception Distance \cite{heusel2017gans}) and binary classification accuracy (defined as whether a specified attribute, or collection of attributes, is present or not in the corresponding generated output image according to a pre-trained classifier). These two metrics are chosen in order to assess each model's ability to match the target distribution from the perspective of both perceptual image quality and visual accuracy. We recognise that FID has limitations as a measure of quality/diversity \cite{jayasumana2024rethinking}, however it remains a standard metric for evaluation of image generation methods \cite{du2019implicit,liu2022compositional,huang2024gan,esser2024scaling}, and serves as a highly informative tool for our purposes.

We conduct all quantitative experiments for 1, 2 and 3 attributes per image for all 3 datasets, totalling 9 quantitative experiments. We use a temperature of $0.9$ when generating samples for our quantitative experiments. Details of how accuracy scores are obtained are in Appendix. Excluded from our experiments are unconditional methods such as R3GAN \cite{huang2024gan} and those with a primary application to text-to-image (MM-DiT \cite{esser2024scaling}).

\begin{table*}[htbp]
    \scriptsize
    \centering
    \caption{Quantitative results (error rate and FID score) on the Positional CLEVR dataset}
    \vspace{0.6em}
    \begin{tabular}{lr@{}lcr@{}lcr@{}lc}
    \toprule
     \multirow{2}{*}[-0pt]{Method} & \multicolumn{3}{c}{\textbf{1 Component}} & \multicolumn{3}{c}{\textbf{2 Components}} & \multicolumn{3}{c}{\textbf{3 Components}} \\
      & \multicolumn{2}{c}{Err (\%) $\downarrow$} & FID $\downarrow$ & \multicolumn{2}{c}{Err (\%) $\downarrow$} & FID $\downarrow$ & \multicolumn{2}{c}{Err (\%) $\downarrow$} & FID $\downarrow$ \\
      \midrule
      StyleGAN2-ADA \cite{karras2020training} & 62.72&$\pm1.37$ & 57.41 & \multicolumn{2}{c}{-} & - & \multicolumn{2}{c}{-} & - \\
      StyleGAN2 \cite{karras2020analyzing} & 98.96&$\pm0.29$ & 51.37 & 99.96&$\pm0.04$ & 23.29  &100.00&$\pm0.00$ &19.01 \\
    LACE \cite{nie2021controllable} & 99.30&$\pm0.24$ & 50.92& 100.00&$\pm0.00$ & 22.83 & 100.00&$\pm0.00$ & 19.62 \\
    GLIDE \cite{nichol2021glide} & 99.14&$\pm0.26$ & 61.68 & 99.94&$\pm0.06$ & 38.26 & 100.00&$\pm0.00$ & 37.18 \\
    EBM \cite{du2020compositional} &  29.46&$\pm1.29$ & 78.63 & 71.78&$\pm1.27$ & 65.45  & 92.66&$\pm0.74$ & 58.33 \\
    Composed GLIDE \cite{liu2022compositional}  & \underline{13.58}&$\pm0.97$ & \underline{29.29} & \underline{40.80}&$\pm1.39$ & \underline{15.94} & \underline{68.64}&$\pm1.31$ & \textbf{10.51} \\
    \midrule
     Ours & \textbf{0.70}&$\pm0.24$& \textbf{13.76} & \textbf{1.82}&$\pm0.38$ & \textbf{15.30} & \textbf{4.96}&$\pm0.61$ & \underline{16.23} \\
     \bottomrule
    \end{tabular}
    \label{tab:quant_clevr_pos}
    \vspace{-0em}
\end{table*}

In comparison to the 6 baseline compositional results reported in \cite{liu2022compositional}, our method \emph{exceeds or matches the accuracy of the previous state-of-the-art in all nine settings}, while attaining highly competitive FID scores across the three datasets. Particularly noteworthy are our accuracy results for Positional CLEVR, for which our method attains an error rate of $0.70\%$, $1.82\%$ and $4.96\%$ on $1$, $2$ and $3$ input components respectively, where the previous state-of-the-art attained error rates of $13.58\%$, $59.20\%$ and $68.64\%$ respectively (Table \ref{tab:quant_clevr_pos}). We see similarly dramatic improvements much harder Relational CLEVR dataset (Table \ref{tab:quant_clevr_rel}) and significant improvements on the FFHQ dataset (Table \ref{tab:quant_ffhq}).

We speculate that the dramatic accuracy improvements offered by our method can be attributed to the fact that the introduction of the discrete representation learning step (VQ-VAE or VQ-GAN) facilitates the emergence of an expressive and richly compositional ``visual language'' to represent images, while the conditional parallel token model offers a strongly regularised and highly calibrated model for the visual language. We conjecture that these effects synergise to produce an efficient, accurate and robust compositional method.

While the FID scores of our method are not the best in all instances (Tables \ref{tab:quant_clevr_pos}, \ref{tab:quant_clevr_rel}, \ref{tab:quant_ffhq}), our method obtains lower FID than the next-best competitor (ranked by error rate) in 7 out of 9 experiments. We recognise that there will always be a trade-off between accuracy and FID; by design, higher concept weights result in stronger/more biased expression of the desired features (Fig. \ref{fig:vary_w_smile}). Nonetheless, scatter plots of error rate against FID indicate that our method empirically lies on the Pareto front in all 9 settings (please see Fig. \ref{fig:intro_scatter} and Appendix for full scatter plots).

\begin{table*}[t]

\caption{Quantitative results (error rate and FID score) on the Relational CLEVR dataset}
\scriptsize
\centering
\begin{tabular}{lr@{}lcr@{}lcr@{}lc}
\toprule
\multirow{2}{*}[-0pt]{Method} & \multicolumn{3}{c}{\textbf{1 Component}} & \multicolumn{3}{c}{\textbf{2 Components}} & \multicolumn{3}{c}{\textbf{3 Components}} \\
& \multicolumn{2}{c}{Err (\%) $\downarrow$} & FID $\downarrow$ & \multicolumn{2}{c}{Err (\%) $\downarrow$} & FID $\downarrow$ & \multicolumn{2}{c}{Err (\%) $\downarrow$} & FID $\downarrow$ \\
\midrule
StyleGAN2-ADA \cite{karras2020training} & \underline{32.29}&$\pm1.32$ & \underline{20.55} & \multicolumn{2}{c}{-} & - & \multicolumn{2}{c}{-} & - \\
StyleGAN2 \cite{karras2020analyzing} & 79.82&$\pm1.14$ & 22.29 & 98.34&$\pm0.36$ & 30.58 & 99.84&$\pm0.11$ & 31.30 \\
LACE \cite{nie2021controllable} & 98.90&$\pm0.30$ & 40.54 & 99.90&$\pm0.09$ & 40.61 & 99.96&$\pm0.04$ & 40.60 \\
GLIDE \cite{nichol2021glide} & 53.80&$\pm1.41$ & \textbf{17.61} & 91.14&$\pm0.80$ & \textbf{28.56} & 98.64&$\pm0.33$ & 40.02 \\
EBM \cite{du2020compositional} & \textbf{21.86}&$\pm1.17$ & 44.41 & \underline{75.84}&$\pm1.21$ & 55.89 & \underline{95.74}&$\pm0.57$ & 58.66 \\
Composed GLIDE \cite{liu2022compositional} & 39.60&$\pm1.38$ & 29.06 & 78.16&$\pm1.17$ & 29.82 & 97.20&$\pm0.47$ & \textbf{26.11} \\
\midrule
Ours  & \textbf{21.84}&$\pm1.17$ & 30.00 & \textbf{56.94}&$\pm1.40$ & \underline{28.87} & \textbf{85.70}&$\pm0.99$ & \underline{30.34} \\
\bottomrule
\end{tabular}
\label{tab:quant_clevr_rel}
\end{table*}

\begin{table*}[t]
    \scriptsize
    \centering
    \caption{Quantitative results (error rate and FID score) on the FFHQ dataset}
  \begin{tabular}{lr@{}lcr@{}lcr@{}lc}
\toprule
\multirow{2}{*}[-0pt]{Method} & \multicolumn{3}{c}{\textbf{1 Component}} & \multicolumn{3}{c}{\textbf{2 Components}} & \multicolumn{3}{c}{\textbf{3 Components}} \\
& \multicolumn{2}{c}{Err (\%) $\downarrow$} & FID $\downarrow$ & \multicolumn{2}{c}{Err (\%) $\downarrow$} & FID $\downarrow$ & \multicolumn{2}{c}{Err (\%) $\downarrow$} & FID $\downarrow$ \\
\midrule
        StyleGAN2-ADA \cite{karras2020training}  & 8.94&$\pm0.81$ & \textbf{10.75} & \multicolumn{2}{c}{-} & - & \multicolumn{2}{c}{-} & - \\
        StyleGAN2 \cite{karras2020analyzing} & 41.10& $\pm1.39$ & \underline{18.04}& 69.32&$\pm1.30$ & \underline{18.06}& 83.04&$\pm1.06$ & \underline{18.06} \\
        LACE \cite{nie2021controllable} & 2.40&$\pm0.43$ & 28.21 & \underline{4.34}&$\pm0.58$ & 36.23 & \underline{19.12}&$\pm1.11$ & 34.64 \\
        GLIDE \cite{nichol2021glide} & 1.34&$\pm0.33$ & 20.30 & 51.32&$\pm1.41$ & 22.69 & 72.76&$\pm1.26$ & 21.98 \\
        EBM \cite{du2020compositional} & 1.26&$\pm0.32$ & 89.95 & 6.90&$\pm0.72$ & 99.64 & 69.99&$\pm1.30$ & 335.70 \\
        Composed GLIDE \cite{liu2022compositional} & \underline{0.74}&$\pm0.24$ & 18.72 & 7.32&$\pm0.74$ & \textbf{17.22} & 31.14&$\pm1.31$ & \textbf{16.95} \\
       \midrule
     Ours & \textbf{0.22}&$\pm0.13$ & 21.52 & \textbf{0.62}&$\pm0.22$ & 28.25 & \textbf{0.82}&$\pm0.26$ & 33.80 \\
     \bottomrule
\end{tabular}
    \label{tab:quant_ffhq}
\end{table*}

\subsection{Qualitative Experiments}
Here we also qualitatively investigate the usefulness of our approach outside of the rigorous quantitative experimental settings. In particular, we investigate the controllability offered by logical conjunction and negation of prompts, as well as the qualitative effect of concept weighting. In addition to the models trained for our quantitative experiments, some of the experiments below apply our method to a pre-trained text-to-image parallel token prediction model (\cite{chang2023muse}). We choose the aMUSEd \cite{patil2024amused} implementation of MUSE because it is publicly accessible (as of the time of writing), in addition to being trained on a large, open dataset \cite{schuhmann2022laion}, which facilitates the open-ended generation which we aim to explore here. Additional qualitative results are in Appendix.

\textbf{Concept negation:} Fig.~\ref{fig:concept_negation} demonstrates the application of concept negation using aMUSEd text-to-image parallel token prediction \cite{patil2024amused}. We compare each example against a single-prompt CFG baseline using the same model. We focus on problematic cases where the underlying text-image model is incapable of properly interpreting negation in the linguistic sense, which is especially pertinent when the concept being negated may be considered an essential characteristic of the concept from which it is being negated (e.g. ``a king \textbf{not} wearing a crown"). Our method allows us to achieve more specific outputs without changing or fine-tuning the underlying model, even in cases where the underlying model fails to comprehend the original negated prompt.

\textbf{Out-of-distribution generation:} We demonstrate our model's ability to generalise to compositions of conditions that are not seen in training. We focus on the (positional) CLEVR dataset, in which individual training samples have at most 5 objects per image. Fig. \ref{fig:out_of_distribution} contains generated samples for input conditions specifying between 6 and 8 objects per image. We make two key observations of  Fig. \ref{fig:out_of_distribution}: (1) that our method successfully generalises outside the distribution of the training data and (2) that re-running the same input gives varied outputs, i.e. the model has not over-fit to always generate the same objects in the same position.

\iffalse

\fi

\textbf{Varying the concept weight:} Fig. \ref{fig:vary_w_smile} illustrates the effect of varying the concept weighting parameter $w$ for a specific input condition (in this case, the weighting of the ``no glasses'' attribute of FFHQ). Keeping other concept weights the same ($w_\textrm{no\_glasses}=w_\textrm{male}=3.0$), we vary $w_\textrm{smile}$ from $-3.0$ to $3.0$.
The outputs in Fig.\ref{fig:vary_w_smile} are consistent with the expectation that the concept weighting method should allow for an interpretable degree of controllability over the generated outputs. Similar visualisations for the other two FFHQ attributes are in Appendix. 

\subsection{Parameter Count and Sampling Time}
Table \ref{tab:param_time_comp} contains a comparison of our method to the most similar methods in the literature (composed EBM \cite{du2020compositional} and composed GLIDE \cite{liu2022compositional}. We compare our method against these methods in particular for 2 reasons: (1) they are compositional and iterative like our own method, and (2) they are generally closest to ours (lowest) in terms of error rate on the three datasets studied. We compare in terms of total parameters and the time taken to generate both a single image and a batch of 25 images on our hardware (NVIDIA GeForce RTX 3090) with 3 input conditions (Positional CLEVR dataset). Runs of baseline methods use the official PyTorch implementations from \cite{liu2022compositional} with default settings (corresponding to the baseline results in Tables \ref{tab:quant_clevr_pos},\ref{tab:quant_clevr_rel} and \ref{tab:quant_ffhq}). The results in Table \ref{tab:param_time_comp} show that our method runs in a fraction of the time of existing approaches while having a comparable number of parameters (and smaller error rate: see Tables \ref{tab:quant_clevr_pos},\ref{tab:quant_clevr_rel}, \ref{tab:quant_ffhq}). Altogether, we see a $2.3\times$ to $12.0\times$ speedup across our speed experiments compared with the baselines.
% TODO: run https://github.com/yilundu/ebm_compositionality with default settings 
\begin{table}[t]
    \scriptsize
    \caption{Parameter counts and sample times for 3 input components}
    \vspace{0.0em}
    \centering
    \begin{tabular}{lccc}
    \toprule
        Method&Parameters&Sample Time&Sample Time\\
        &&Per Image&Per Batch of 25\\
    \midrule
        EBM & 33 mil & 5.99s$\pm$0.17 & 108.57s$\pm$0.93 \\
        Composed GLIDE & 385 mil & 4.92s$\pm0.17$ & 73.92s$\pm$0.70\\
    \midrule
        Ours & 108 mil & \textbf{2.11s}$\pm$0.39 & \textbf{9.08s}$\pm$0.39\\
    \bottomrule
    \end{tabular}
    \label{tab:param_time_comp}
    \vspace{-0em}
\end{table}

\iffalse
\subsection{Interpretability of the negation operator}
In order to interrogate the interpretability of our approach, we conduct one additional experiment, focusing on the FFHQ (faces) dataset. In particular, we aim to investigate how well our \textit{negation} operator generalises for binary attributes. We evaluate this by computing the correlation coefficient $\rho$ between compositional log-probability scores (first iteration only) and their ``negated complement'' for each binary attribute of FFHQ (e.g. we compute the correlation of log-probs between "female" and "not male"). In the ideal generalisation scenario, such ``complement pairs'' should be perfectly correlated ($\rho$=1). To illustrate, we visualise the correlations for 3 such pairs using scatter plots in Fig.~\ref{fig:negated-scatters}. We find that the overall correlation is are ??, indicating ??.
\fi

%% file: cvpr-2026/sec/5_discussion.tex
\section{Discussion and Limitations}
\label{sec:discussion}
Through varied quantitative and qualitative experiments, we have demonstrated that our formulation for compositional generation with iterative sampling methods is readily applicable to a range of tasks for both newly trained and out-of-the-box pre-trained models. We demonstrated state-of-the-art performance in terms of the error rate of the generated results, in addition to obtaining competitive sample quality as measured by FID scores. This is achieved with minimal extra cost in terms of memory, since only the accumulated log-probability outputs need to be retained at inference time. The simplicity of our method offers further advantages, including ease of implementation (facilitating integration with existing discrete generation pipelines) in addition to improved interpretability, since the composition operator can be thought of as directly taking the ``intersection'' between two discrete distributions.

The strong quantitative metrics of our method are complemented by its \textit{out-of-distribution} generation capability and \textit{controllability}. The significance of the results of our quantitative experiments is further reinforced by the fact that we used the same experimental settings for all three of the datasets studied, without extensive fine-tuning of hyperparameters, training runs or model architecture. Our method further provides a $2.3\times$ to $12\times$ speedup over comparable approaches on our hardware.

Similarly to compositional methods for continuous processes \cite{du2020compositional,liu2022compositional}, our method requires $(n+1)$ times the number of feed-forward operations compared to standard iterative approaches of the same architecture, where $n$ is the number of conditions imposed on the output. This is a direct consequence of the mathematical formulation of the approach, however this is largely mitigated by the fact that our method can produce accurate and high-quality outputs in only a small number of iterations \cite{du2020compositional,liu2022compositional}.

Our method makes the necessary assumption that input conditions are conditionally independent given $\boldsymbol{x}$, i.e. $P(\boldsymbol{c}_1,\boldsymbol{c}_2|\boldsymbol{x})\propto P(\boldsymbol{c}_1|\boldsymbol{x})P(\boldsymbol{c}_2|\boldsymbol{x})$ for all condition pairs $\boldsymbol{c}_1,\boldsymbol{c}_2$. It is possible in practical scenarios that this underlying assumption of our approach does not hold, for example due to biases in the training data. The importance-weighting capability of our method can mitigate this in part by allowing the user to compensate for potential biases. However, we speculate that greater robustness would be better achieved through an unbiased backbone model. Training unbiased models for image generation is beyond the scope of this work, and remains an open challenge, especially in the context of text-to-image generation \cite{wan2024survey}. Further to this, we have not yet explored principled methods for choosing the condition weighting coefficients $w_i$, which may be an interesting direction for future work (e.g. producing a learned concept-weighting policy).

%% file: cvpr-2026/sec/6_conclusion.tex
\section{Conclusion}
We have proposed a novel method for enabling precisely controllable conditional image generation by composing discrete iterative generative processes. The empirical success of our method across the axes of sampling speed, error rate and FID demonstrates a conceptual step beyond the previous state-of-the-art for compositional generation. We further show that our approach can be applied to an out-of-the-box pre-trained text-to-image model to allow for principled and controllable generation without any fine-tuning. The prospect of applying our method outside of image generation (such as multi-prompt text generation) remains an intriguing possibility for future work. Altogether, we believe our work provides a strong foundation for future work in the direction of controllable generation in discrete spaces.